\documentclass[twoside,11pt]{article}

\usepackage{jmlr2e}
\usepackage{amsmath}  
\usepackage{amssymb}  
\usepackage{multirow}
\usepackage{listings}
\usepackage{graphicx} 
\usepackage{subcaption}
\usepackage{fancyhdr}
\usepackage{titlesec}
\usepackage{titling}
\titleformat{\section}  
          {\normalfont\LARGE\bfseries}{\thesection}{1em}{}

\begin{document}
\title{LuminanceL1Loss: A loss function which measures percieved brightness and colour differences}

\author{Dominic De Jonge}

\maketitle

\begin{abstract}
We introduce LuminanceL1Loss, a novel loss function designed to enhance the performance of image restoration tasks. We demonstrate its superiority over MSE when applied to the Retinexformer, BUIFD and DnCNN architectures. Our proposed LuminanceL1Loss leverages a unique approach by transforming images into grayscale and subsequently computing the MSE loss for both grayscale and color channels. Experimental results demonstrate that this innovative loss function consistently outperforms traditional methods, showcasing its potential in image denoising and other related tasks in image reconstruction. It demonstrates gains up to 4.7dB. The results presented in this study highlight the efficacy of LuminanceL1Loss for various image restoration tasks.
\end{abstract}

\section{Introduction}
Image denoising is a critical challenge in computer vision with implications for diverse domains including autonomous vehicles, surveillance, and remote sensing. This task involves restoring visual fidelity degraded by noise in the imaging process. Conventionally, L1 and L2 losses have been the predominant metrics for optimizing disparity between predicted and ground truth images. However, limitations capturing the intricacies of noisy images have motivated developing alternative losses to improve denoising, and images with identical MSE can have vastly different percieved quality[\cite{1284395}]. 

In response, we propose LuminanceL1Loss, a loss function addressing shortcomings of existing losses for image denoising. Our rationale is the ability to incorporate both grayscale and color during calculation. This unique approach accounts for perceived brightness, enabling a more comprehensive perspective to optimize denoising. Incorporating grayscale provides insights into luminance, improving restoration of noisy images (as measured by PNSR and SSIM).

We test this loss function on the BSD68 color dataset[\cite{MartinFTM01}] for the DnCNN[\cite{dncnn}] and BUIFD[\cite{BUIFD}] models, as well as the LOL-v1[\cite{lol1}], LOL-v2 real and LOL-v2 synthetic datasets[\cite{lol2}] for the Retinexformer model[\cite{retinexformer}]. This is because they are quoted in the original papers and therefore this is representative of the best-case scenario for the architectures. 

This paper is structured as follows: Section II surveys related work, emphasizing existing techniques and gaps addressed. Section III elaborates on LuminanceL1Loss, explaining the design and mathematical formulation. Section IV details the experimental methodology including data, training, and evaluation. Section V presents results and discussion, highlighting performance across diverse denoising scenarios. Finally, Section VI concludes with findings, contributions, and future opportunities.

\section{Related works}

\subsection{Retinexformer}

Retinexformer [\cite{retinexformer}] is a Transformer-based method for low-light image enhancement. It formulates a One-Stage Retinex-based Framework (ORF) to model corruptions and uses an Illumination-Guided Transformer (IGT) as the corruption restorer.

Specifically, ORF contains an illumination estimator to output a light-up map and enhance visibility. It also has a corruption restorer to suppress noise, artifacts, under-/over-exposure and color distortion in a single feedforward pass.

The key innovation in Retinexformer is the IGT corruption restorer. It employs a novel Illumination-Guided Multi-head Self-Attention (IG-MSA) mechanism. IG-MSA leverages illumination representations from ORF to guide modeling of long-range dependencies between image regions. This allows IGT to effectively capture non-local interactions for enhancement.

Compared to prior works, Retinexformer achieves significantly higher PSNR and better perceptual quality. It improves state-of-the-art by 1-6 dB on various benchmarks. Retinexformer also demonstrates greater efficiency than other Transformer architectures through the proposed IG-MSA.

In summary, Retinexformer explores the potential of Transformers for low-light image enhancement by designing an end-to-end architecture incorporating domain knowledge. It sets a new benchmark for the task. Our method is partly inspired by the success and novelty of Retinexformer. However, we introduce additional innovations as described in later sections.

\subsection{DnCNN}

A deep convolutional neural network architecture called DnCNN[\cite{dncnn}] is specifically targeted for image denoising. DnCNN adopts a residual learning formulation, where the network learns to predict the noise component rather than the clean image directly. This allows the model to focus on removing the latent clean image in the hidden layers. DnCNN integrates residual learning with batch normalization to speed up training and boost denoising performance.

Compared to prior discriminative learning models like MLP[\cite{6247952}] and TNRD[\cite{7527621}] that train separate models for each noise level, DnCNN can handle blind Gaussian denoising across a wide range of noise levels using a single model. DnCNN also extends this framework to other image restoration tasks such as super-resolution and JPEG deblocking.

Experiments demonstrate superior quantitative and qualitative performance to state-of-the-art methods like BM3D and TNRD for Gaussian denoising. DnCNN is also efficient to implement on GPUs. The effectiveness of DnCNN for various image denoising problems highlights the potential of deep convolutional neural networks and residual learning for image reconstruction.

\subsection{BUIFD}

Blind Universal Image Fusion Denoising (BUIFD)[\cite{BUIFD}]. It is a deep learning approach for image denoising that achieves state-of-the-art performance. A key innovation of BUIFD is its ability to handle a wide range of noise levels in a blind, universal manner using a single model.

The authors derive an optimal Bayesian denoising solution under assumptions of additive Gaussian noise and a Gaussian image prior. This theoretical fusion function combines a learned image prior with the noisy input based on a predicted signal-to-noise ratio.

While real images do not precisely follow a Gaussian prior, BUIFD adapts this architecture by disentangling the feature space. Separate branches predict the image prior, noise level, and final fused output. An auxiliary loss on noise level prediction helps the model generalize to unseen noise conditions.

Experiments on standard benchmarks like BSD68 demonstrate BUIFD's state-of-the-art denoising performance. The method shows significant improvements in PSNR and SSIM over a wide range of noise levels, outperforming previous approaches. Critically, BUIFD generalizes much better to unseen noise levels compared to earlier methods.

In conclusion, BUIFD advances the state-of-the-art in blind universal image denoising through an interpretable architecture inspired by Bayesian principles. The impressive results highlight the potential of theoretically-grounded deep learning models for image restoration.

\subsection{Other loss functions}

L1 and L2 loss are the most common loss functions currently used in image reconstruction, and are generally effective, however they have known issues. One issue exhibited by L2 loss, for example, is that it can compare images put through the same compression, but different artefact types are not treated equally or in a way informed by human vision[\cite{article}].

\[\mathcal{L}_1 = \frac{1}{N} \sum_{i=1}^{N} |y_i - \hat{y}_i|\]

\[\mathcal{L}_2 = \frac{1}{N} \sum_{i=1}^{N} (y_i - \hat{y}_i)^2\]

Perceptual losses originated from research on image super-resolution and style transfer. This can more meaningfully capture the features in the input and output images[\cite{cxloss}], and is therefore better than L1 and L2 loss. Percertual loss involves calculating the difference between features extracted from a predicted and ground truth image. The features are extracted by passing them through a pre-trained, frozen model.

\[\mathcal{L}_{\text{perceptual}} = \sum_{i=1}^{N} \lambda_i \cdot \|\Phi_i(y) - \Phi_i(\hat{y})\|_2\]

SSIM[\cite{1284395}] is an image quality assessment metric used to measure the perceptual similarity between two images. It compares local patterns of pixel intensities that have been normalized for luminance and contrast.
The structural information is captured using the statistics of small windows in the images. The overall SSIM score between two images is computed as the average of SSIM scores from multiple windows. SSIM ranges from -1 to 1, with higher values indicating greater structural similarity. A value of 1 indicates perfect similarity.

SSIM is perceptually more meaningful than other common metrics because it compares local patterns and structure, rather than just pixel-level differences. It also accounts for inter-dependency of luminance and contrast, and it is sensitive to small geometric distortions or noise. However it is highly dependant on window size selection affects performance. It is also mainly valid for low levels of distortion.

\[\text{SSIM}(x, y) = \frac{{(2\mu_x\mu_y + C_1)(2\sigma_{xy} + C_2)}}{{(\mu_x^2 + \mu_y^2 + C_1)(\sigma_x^2 + \sigma_y^2 + C_2)}}\]

\section{LuminanceL1Loss}

\subsection{Motivation}

There have been very few developments in recent years relating to image restoration loss functions. As such, most state of the art models use L2 loss, for example we can see it used in the reference implementation of BUFID[\cite{BUIFD}]. We realised that no loss functions take the perceived brightness of an image into account when calculating the similarity of 2 images, meaning 2 images with the same L2 loss can look very different  [\cite{1284395}]. Methods that reduce this effect, like Perceptual loss using pretrained Computer Vision models, are often far more computationally expensive[\cite{mustafa2021training}] and therefore are not used. We propose a new loss function, LuminanceL1Loss to allow for the brightness of an image to be used in the loss function when training an image reconstruction model. 

\subsection{Pixel-level component}

In the process of designing an effective loss function for our image denoising model, we chose the L1 loss for the pixel-level component. This selection was based on practical considerations, empirical evidence, and the need for comparative analysis. The L1 loss, or Mean Absolute Error (MAE), is known for its computational simplicity. This allows for faster training when compared to Perceptual losses using ViT[\cite{dosovitskiy2021image}] or other similar pretrained vision models. 

The L1 loss also has a track record of effectiveness in various state-of-the-art (SOTA) models, including those used in image denoising. Its simplicity and ability to capture pixel-level differences have contributed to its popularity in the machine learning community. By adopting this loss function, we align with established practices in the field.

Furthermore, the use of the L1 loss enables straightforward comparisons with existing denoising methods. Image denoising algorithms often struggle to distinguish between moisy and noise-free regions. Employing a well-established loss function allows for more direct comparisons, particularly when assessing the impact of luminance being included in the loss. This facilitates a rigorous evaluation of our approach in the context of existing research.

In summary, our choice of the L1 loss for the pixel-level component of our loss function is grounded in considerations of computational efficiency, empirical evidence, and the need for objective comparisons with prior work in image denoising. This decision aims to streamline our model's performance and position it as a reliable contender in the ongoing development of denoising techniques.

\subsection{Brightness component}

In our proposed method, we utilize an additional loss term to capture luminance differences between the prediction and ground truth images. Specifically, we first convert both the predicted image and ground truth image to grayscale versions, giving us two single channel images that purely contain luminance information. We then calculate the L1 loss between these grayscale images.

Taking the L1 norm allows us to quantify the absolute difference in luminance values per pixel between the prediction and ground truth. Minimizing this luminance loss during training will encourage the model to not only match color information, but also better reproduce the underlying lighting and shading characteristics present in the ground truth image. The network learns to pay attention to modeling lighting and shading patterns, not just matching colors. Adding this extra supervision signal on the achromatic domain further regularizes the model and improves generalization. It is defined here:

\[\mathcal{L}_{grey} = |y_i \cdot([0.2989, 0.5870, 0.1140]) - \hat{y}_i \cdot([0.2989, 0.5870, 0.1140])|\]

\subsection{Loss function}

We combine the brightness and L1 loss linearly:

\[\mathcal{L}_{total} = |y_i - \hat{y}_i| + \lambda|y_i \cdot([0.2989, 0.5870, 0.1140]) - \hat{y}_i \cdot([0.2989, 0.5870, 0.1140])|\]

Where $\lambda$ is a constant.

\section{Experiments}

We tested out loss on 3 models across 4 datasets. We used the PNSR and SSIM metrics when evaluating RetinexFormer on LOL-v1[\cite{lol1}], LOL-v2 real and LOL-v2 synthetic[\cite{lol2}]. As well as this, we used PNSR on the Color BSD68 dataset[\cite{MartinFTM01}] to test the CDnCNN[\cite{dncnn}] and CBUIFD[\cite{BUIFD}] models.

\subsection{Low-Light Image Enhancement}

We tested the Retinexformer architecture on three low light image enhancement datasets used by the original authors of RetinexFormer: LOL-v1, LOL-v2 synthetic and LOL-v2 real. We used these datasets because they show the Retinexformer model performing well. We used the reference implementation (\href{https://github.com/caiyuanhao1998/Retinexformer}{https://github.com/caiyuanhao1998/Retinexformer}) for all of these, keeping all settings the same except for the loss function. Our goal was therefore to analyze if LuminanceL1Loss improves the  low-light enhancement ability of Retinexformer over a baseline of MSE, as well as other models that the original authors used for comparison, to see how much worse this loss would perform. 
\subsubsection{Quantitative results}

\begin{table*}
\centering
\begin{tabular}{ |p{3cm}||p{3cm}|p{3cm}|p{3cm}|p{3cm}| }
 \hline
  Model& Complexity & LOL-v1&LOL-v2-syn&LOL-v2-real\\
      & FLOPS (G) / params (M) & PNSR / SSIM & PNSR / SSIM & PNSR / SSIM\\
 \hline
SID [\cite{Chen_2019_ICCV}] & 13.73 / 7.76 & 14.35 / 0.436 & 15.04 / 0.610 & 13.24 / 0.442 \\
3DLUT [\cite{Zeng_2020}] & 0.075 / 0.59  & 14.35 / 0.445 & 18.04 / 0.800 & 17.59/  0.721 \\
DeepUPE [\cite{8953588}] & 21.10 / 1.02 & 14.38/  0.446 & 15.08 / 0.623 & 13.27 / 0.452 \\
RF [\cite{Kosugi_Yamasaki_2020}] & 46.23 / 21.54 & 15.23 / 0.452 & 15.97 / 0.632 & 14.05 / 0.458 \\
DeepLPF [\cite{Moran_2020_CVPR}] & 5.86 / 1.77 & 15.28/  0.473 & 16.02 / 0.587 & 14.10 / 0.480 \\
IPT [\cite{chen2021pretrained}]& 6887 / 115.31 & 16.27 / 0.504 & 18.30 / 0.811 & 19.80 / 0.813 \\
UFormer [\cite{Wang_2022_CVPR}]& 12.00 / 5.29 & 16.36 / 0.771 & 19.66 / 0.871 & 18.82 / 0.771 \\
RetinexNet [\cite{lol1}]& 587.47 / 0.84 & 16.77 / 0.560 & 17.13 0.798 & 15.47 / 0.567 \\
Sparse [\cite{9328179}]& 53.26 / 2.33 & 17.20 / 0.640 & 22.05 / 0.905 & 20.06 / 0.816 \\
EnGAN [\cite{jiang2021enlightengan}]& 61.01 / 114.35 & 17.48 / 0.650 & 16.57 / 0.734 & 18.23 / 0.617 \\
RUAS [\cite{liu2020retinexinspired}]& 0.83 / 0.003 & 18.23 / 0.720 & 16.55 / 0.652 & 18.37 / 0.723 \\
FIDE [\cite{9156446}]& 28.51 / 8.62 & 18.27 / 0.665 & 15.20 / 0.612 & 16.85 / 0.678 \\
DRBN [\cite{9369069}]& 48.61 / 5.27 & 20.13 / 0.830 & 23.22 / 0.927 & 20.29 / 0.831 \\
KinD [\cite{zhang2019kindling}]& 34.99 / 8.02 & 20.86 / 0.790 & 13.29 / 0.578 & 14.74 / 0.641 \\
Restormer [\cite{Zamir2021Restormer}]& 144.25 / 26.13 & 22.43 / 0.823 & 21.41 / 0.830 & 19.94 / 0.827 \\
MIRNet [\cite{efa0de47e5b44fb0bfc5986dbb40751e}]& 785 / 31.76 & 24.14 / 0.830 & 21.94 / 0.876 & 20.02 / 0.820 \\
SNR-Net [\cite{9878461}]& 26.35 / 4.01 & 24.61 / 0.842 & 24.14 / 0.928 & 21.48 / 0.849 \\
Retinexformer (base)& 15.57 / 1.61 & 25.16 / 0.845 & 25.67 / 0.930 & 22.80 / 0.840\\
\hline
Retinexformer (ours)& 15.57 / 1.61 & 25.30 / 0.841  & 25.98 / 0.931 & 27.50 / 0.875 \\
\hline

\end{tabular}
\caption{PNSR and SSIM on Low-Light Image Enhancement datasets}
\end{table*}

We observed modest PSNR improvements of 0.14dB on LOL-v1 and 0.31dB on LOL-v2 synthetic using our new loss, with negligible impact on SSIM (within 0.005). However, on the LOL-v2 real dataset, we achieved a large 4.7dB PSNR gain and 0.035 higher SSIM over the baseline Retinexformer. This result demonstrates that LuminanceL1Loss provides superior performance over MSE for low-light image enhancement while adding minimal computational overhead (training time was within 5\% of the training time of the MSE model). The results validate that combining MSE and L1 loss on luminance is an effective strategy for improving low-light enhancement of images.

\begin{figure}[ht] 
   \begin{subfigure}{0.3\textwidth}
       \includegraphics[width=\linewidth]{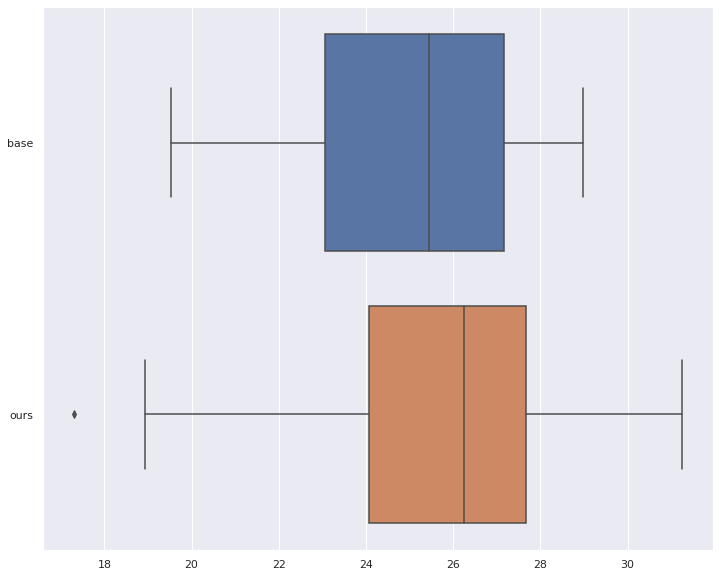}
       \caption{PNSR}
       \label{fig:subim1}
   \end{subfigure}
\hfill 
   \begin{subfigure}{0.3\textwidth}
       \includegraphics[width=\linewidth]{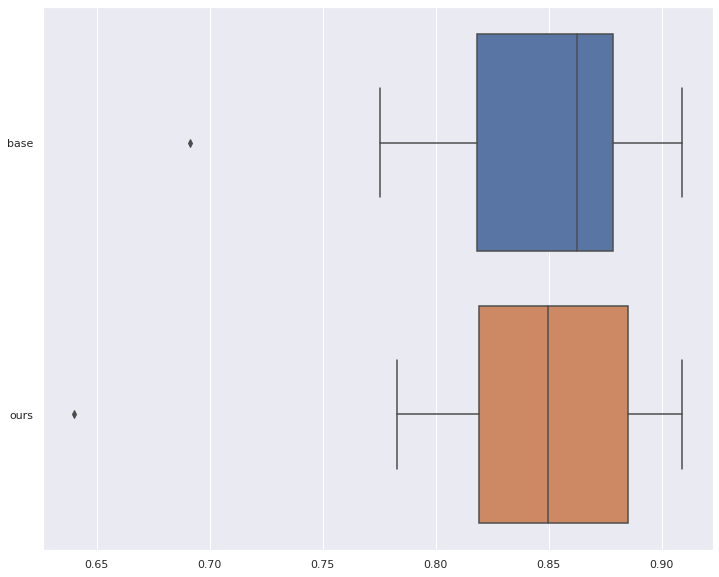}
       \caption{SSIM}
       \label{fig:subim2}
   \end{subfigure}
   \caption{LOL-v1 performance}
   \label{fig:image2}
\end{figure}

In the LOL-V1 dataset, it is clear that our RetinexFormer model has higher PNSR, however this average is dragged down by an outlier. This is also to blame for the poor SSIM performance which otherwise has higher quartiles and would therefore have a higher mean.

\begin{figure}[ht] 
   \begin{subfigure}{0.3\textwidth}
       \includegraphics[width=\linewidth]{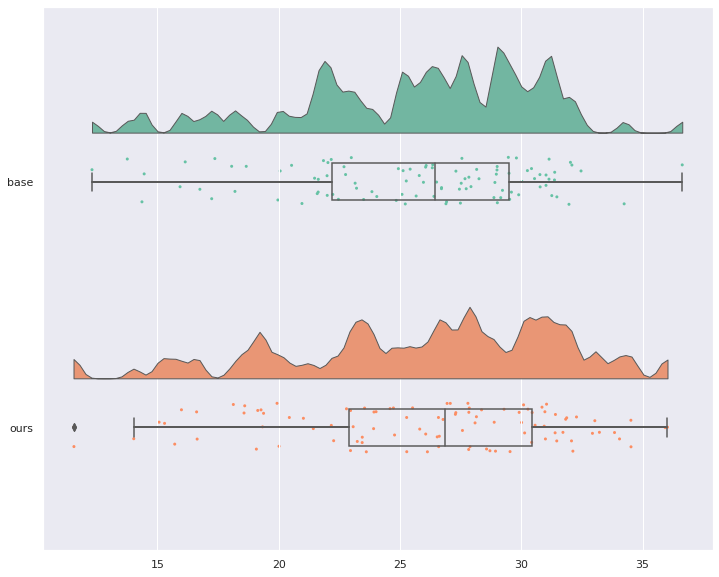}
       \caption{PNSR}
       \label{fig:subim1}
   \end{subfigure}
\hfill 
   \begin{subfigure}{0.3\textwidth}
       \includegraphics[width=\linewidth]{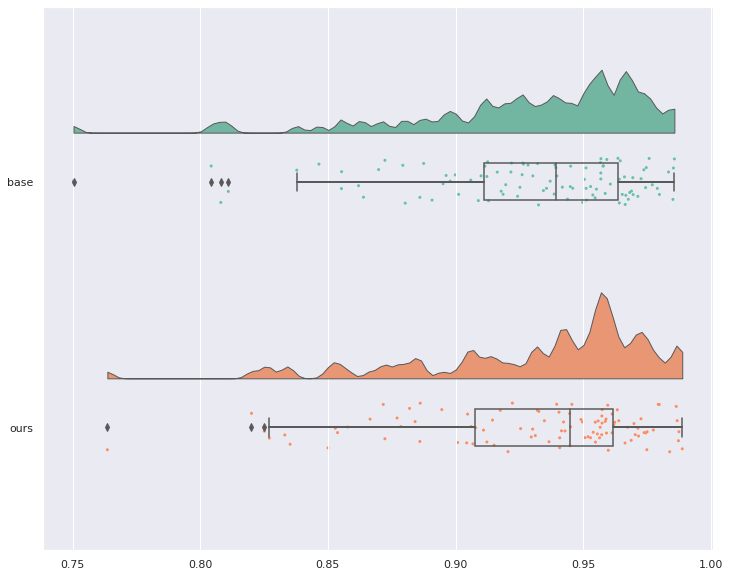}
       \caption{SSIM}
       \label{fig:subim2}
   \end{subfigure}
   \caption{LOL-v2 synthetic performance}
   \label{fig:image2}
\end{figure}

In the LOL-v2 synthetic dataset, the results have much larger spread. As a result, whilst performance gains may seem modest, gains are very large on certain scenes. This is especially noticable on the SSIM score, which has a large spike towards the upper quartile on our model.

\begin{figure}[ht] 
   \begin{subfigure}{0.3\textwidth}
       \includegraphics[width=\linewidth]{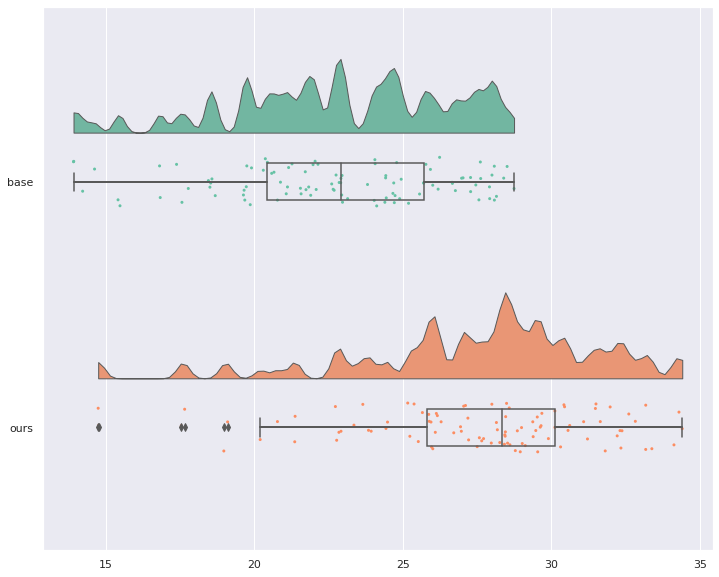}
       \caption{PNSR}
       \label{fig:subim1}
   \end{subfigure}
\hfill 
   \begin{subfigure}{0.3\textwidth}
       \includegraphics[width=\linewidth]{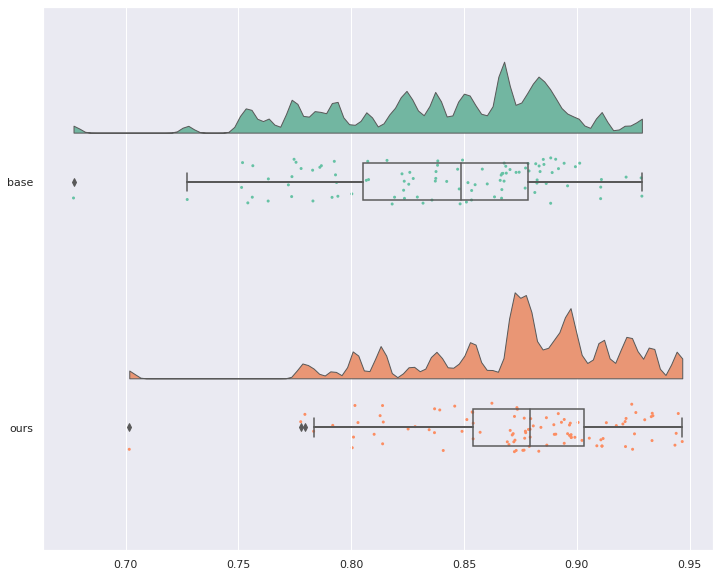}
       \caption{SSIM}
       \label{fig:subim2}
   \end{subfigure}
   \caption{LOL-v2 real performance}
   \label{fig:image2}
\end{figure}

In the LOL-v2 real dataset, our model has much greater and more consistent performance. Whilst there are outliers, there are massive spikes in frequency. This shows that the model is more resilient and generally better, rather than improving on a small amount of scenes.

\subsubsection{Qualitative results}

Here we have chosen 3 images from the LOL-v2 real dataset to compare the performance of Retinexformer trained with MSE to Retinexformer trained with LuminanceL1Loss. These images were chosen to demonstrate different areas where both models struggles, such as scenes involving brigh lights, dark areas and small details.

\begin{figure}[ht] 
   \begin{subfigure}{0.243\textwidth}
       \includegraphics[width=\linewidth]{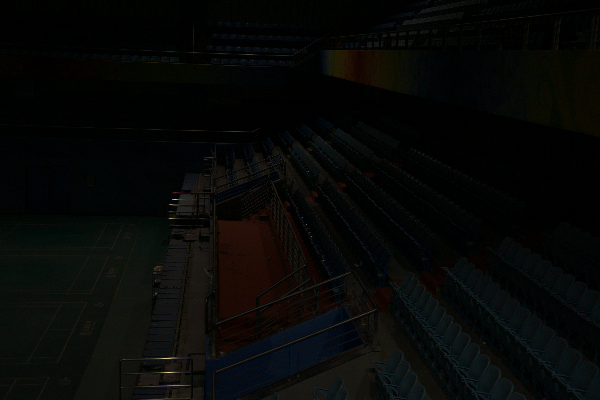}
       \caption{Input}
       \label{fig:subim1}
   \end{subfigure}
\hfill 
   \begin{subfigure}{0.243\textwidth}
       \includegraphics[width=\linewidth]{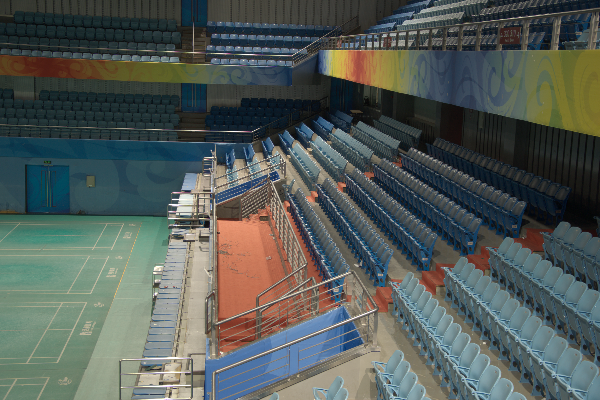}
       \caption{ground truth}
       \label{fig:subim2}
   \end{subfigure}
\hfill 
   \begin{subfigure}{0.243\textwidth}
       \includegraphics[width=\linewidth]{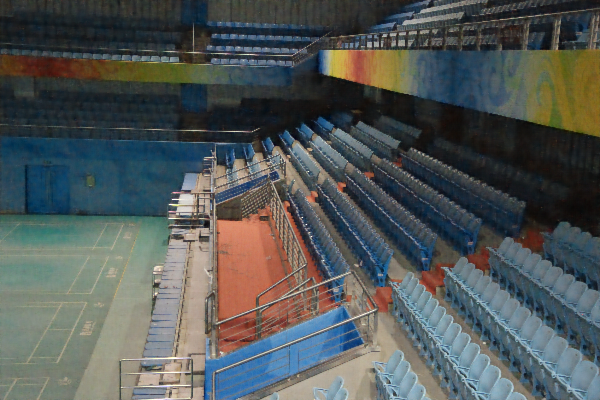}
       \caption{Our Retinexformer}
       \label{fig:subim3}
   \end{subfigure}
\hfill 
   \begin{subfigure}{0.243\textwidth}
       \includegraphics[width=\linewidth]{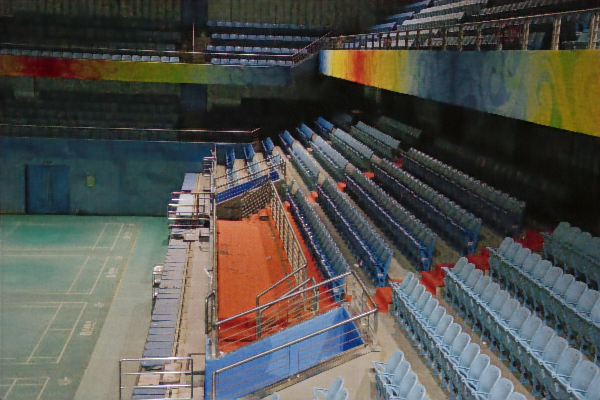}
       \caption{Retinexformer}
       \label{fig:subim3}
   \end{subfigure}

   \caption{Low-light image enhancement image 1}
   \label{fig:image2}
\end{figure}

Figure 4 shows a scenario with many colours. Retinexformer trained with MSE is shown to produce oversaturated results, where as our model has more visually accurate colours, such as the red platform in the middle of the image.

\begin{figure}[ht] 
   \begin{subfigure}{0.243\textwidth}
       \includegraphics[width=\linewidth]{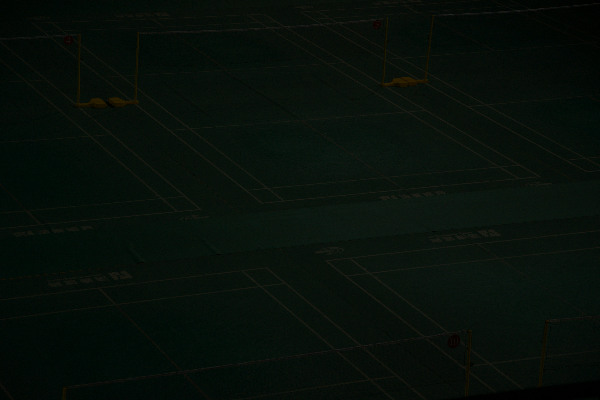}
       \caption{Input}
       \label{fig:subim1}
   \end{subfigure}
\hfill 
   \begin{subfigure}{0.243\textwidth}
       \includegraphics[width=\linewidth]{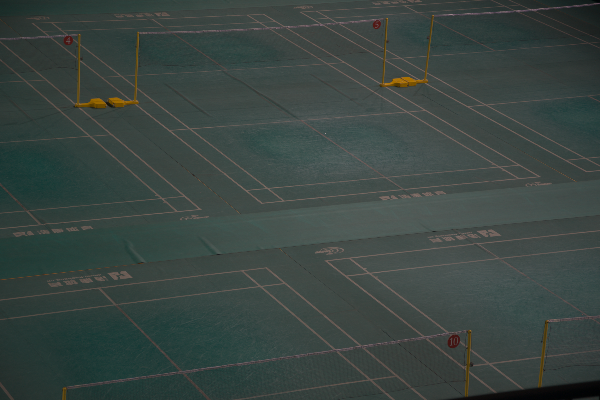}
       \caption{ground truth}
       \label{fig:subim2}
   \end{subfigure}
\hfill 
   \begin{subfigure}{0.243\textwidth}
       \includegraphics[width=\linewidth]{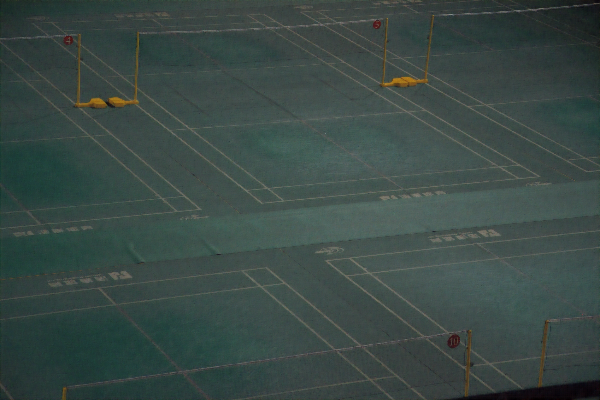}
       \caption{Our Retinexformer}
       \label{fig:subim3}
   \end{subfigure}
\hfill 
   \begin{subfigure}{0.243\textwidth}
       \includegraphics[width=\linewidth]{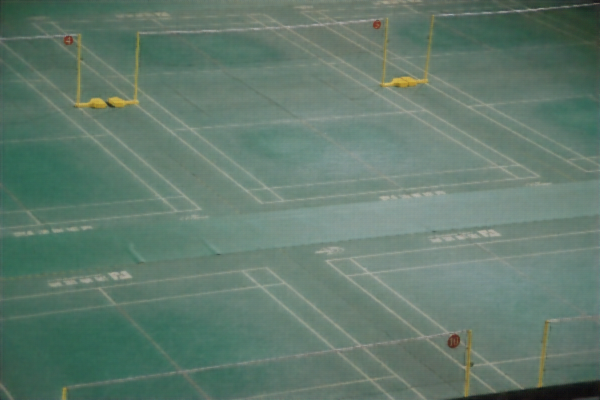}
       \caption{Retinexformer}
       \label{fig:subim3}
   \end{subfigure}

   \caption{Low-light image enhancement image 2}
   \label{fig:image2}
\end{figure}

Figure 5 demonstrates a scene with very few visual queues as to the brightness. Retinexformer trained with MSE is shown to produce bright and washed out results, where as our model has more visually accurate colours.

\begin{figure}[ht] 
   \begin{subfigure}{0.243\textwidth}
       \includegraphics[width=\linewidth]{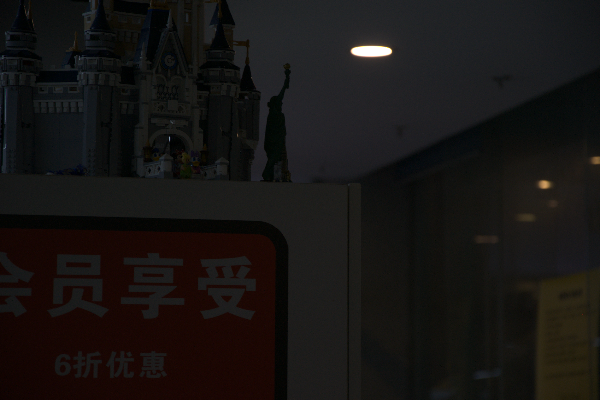}
       \caption{Input}
       \label{fig:subim1}
   \end{subfigure}
\hfill 
   \begin{subfigure}{0.243\textwidth}
       \includegraphics[width=\linewidth]{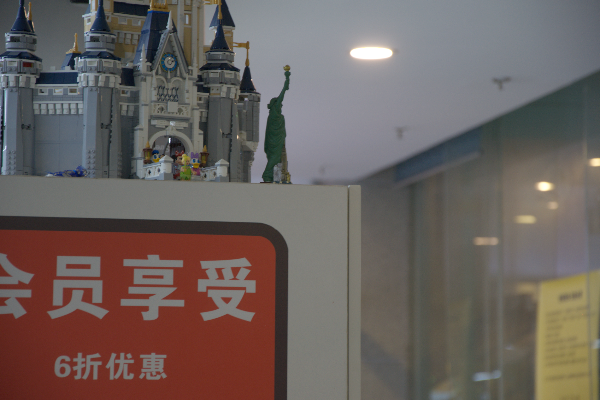}
       \caption{ground truth}
       \label{fig:subim2}
   \end{subfigure}
\hfill 
   \begin{subfigure}{0.243\textwidth}
       \includegraphics[width=\linewidth]{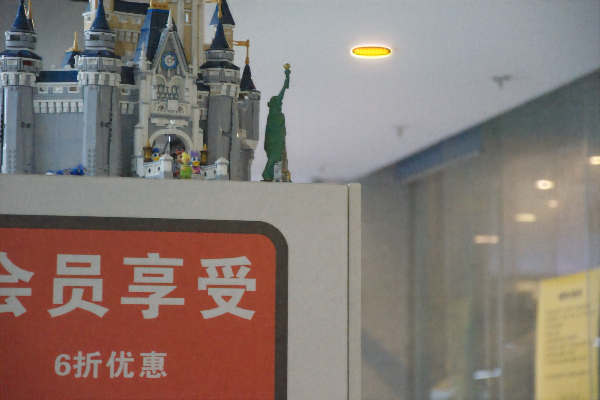}
       \caption{Our Retinexformer}
       \label{fig:subim3}
   \end{subfigure}
\hfill 
   \begin{subfigure}{0.243\textwidth}
       \includegraphics[width=\linewidth]{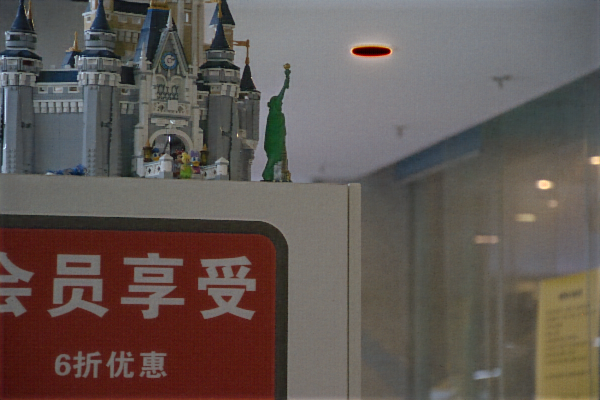}
       \caption{Retinexformer}
       \label{fig:subim3}
   \end{subfigure}

   \caption{Low-light image enhancement image 3}
   \label{fig:image2}
\end{figure}

Figure 6 demonstrates our method's performance in lighting again. When LuminanceL1Loss is used, the sign in the front of the scene is the correct colour. This is not the case for MSE loss. As well as this, the light is a far better, as it is just black in the original work.

\subsection{Blind color image denoising}

We tested blind color image denoising on the color BSD68 dataset. This compared 2 models trained at 2 different levels of noise. We then evaluated these models on 15 levels of noise and the models trained with LuminanceL1Loss to the ones not using it. Whilst many models have achieved far higher PNSR scores than ours, these were not trained blind. This means that our models have no information on how much noise was added (where others would be given this information), which is more realistic for real world applications.

\begin{table}
    \centering
\begin{tabular}{ |p{3cm}||p{3cm}|p{3cm}|p{3cm}|p{3cm}| }
 \hline
  Noise Level& $CDnCNN_{55}$ & $CDnCNN_{55}$ & $CDnCNN_{75}$ & $CDnCNN_{75}$ \\
    (std dev)  & (base) & (ours) & (base) & (ours) \\
 \hline
5 & 40.05  & 40.26 & 39.75 & 40.06 \\
10 & 35.92 & 36.05 & 35.74 & 35.88 \\
15 & 33.57 & 33.68 & 33.46 & 33.53 \\
20 & 31.93 & 32.03 & 31.86 & 31.89 \\
25 & 30.66 & 30.76 & 30.61 & 30.62 \\
30 & 29.61 & 29.71 & 29.59 & 29.59 \\
35 & 28.71 & 28.81 & 28.70 & 28.71 \\
40 & 27.92 & 28.01 & 27.92 & 27.95 \\
45 & 27.16 & 27.28 & 27.19 & 27.24 \\
50 & 26.49 & 26.61 & 26.52 & 26.59 \\
55 & 25.84 & 25.97 & 25.89 & 25.97 \\
60 & 25.23 & 25.36 & 25.27 & 25.37 \\
65 & 24.65 & 24.75 & 24.69 & 24.81 \\
70 & 24.09 & 24.16 & 24.13 & 24.25 \\
75 & 23.52 & 23.64 & 23.59 & 23.72 \\


\hline
\end{tabular}
\caption{PNSR of CDnCNN models on Color BSD68 dataset}
\end{table}
\subsubsection{CDnCNN}

For the CBSD68 dataset using the CDnCNN model, we see small but consistent PNSR gains when using LuminanceL1Loss compared to when not using it. These were greater when the standard deviation used in training was greater. Gains were most consistent and largest when the amount of noise was lowest, with an improvement of over 0.3 dB when the noise had a standard deviation of 5. However, PNSR dropped quickly with higher noise, but at both extremes it is much better than the base models. Overall, using LuminanceL1Loss demonstrates an improvement over a baseline of L1Loss. We used the reference implementation found at \href{https://github.com/cszn/DnCNN}{https://github.com/cszn/DnCNN}

\begin{table}
    \centering

\begin{tabular}{ |p{3cm}||p{3cm}|p{3cm}|p{3cm}|p{3cm}| }
 \hline
  Noise Level& $CBUIFD_{55}$ & $CBUIFD_{55}$ & $CBUIFD_{75}$ & $CBUIFD_{75}$ \\
    (std dev)  & (base) & (ours) & (base) & (ours) \\
 \hline
5 &  40.07 & 40.22 & 40.05 & 40.21 \\
10 & 36.01 & 36.05 & 35.98 & 36.04 \\
15 & 33.66 & 33.68 & 33.65 & 33.68 \\
20 & 32.02 & 32.03 & 32.03 & 32.03 \\
25 & 30.75 & 30.76 & 30.76 & 30.76 \\
30 & 29.72 & 29.71 & 29.71 & 29.72 \\
35 & 28.81 & 28.81 & 28.81 & 28.83 \\
40 & 28.01 & 28.01 & 28.01 & 28.03 \\
45 & 27.27 & 27.28 & 27.28 & 27.30 \\
50 & 26.59 & 26.61 & 26.69 & 26.62 \\
55 & 25.94 & 25.97 & 25.96 & 25.98 \\
60 & 25.33 & 25.36 & 25.34 & 25.36 \\
65 & 24.75 & 24.75 & 24.76 & 24.78 \\
70 & 24.18 & 24.12 & 24.18 & 24.22 \\
75 & 23.62 & 23.44 & 23.64 & 23.68 \\
\hline
\end{tabular}
\caption{PNSR of CBUIFD models on Color BSD68 dataset}
\end{table}

\subsubsection{CBUIFD}
For the CBSD68 dataset using the CBUIFD model, we see small but consistent PNSR gains when using LuminanceL1Loss compared to when not using it. These were greater when the standard deviation used in training was higher. Gains were most consistent and largest when the amount of noise was lowest, with an improvement of over 0.15 dB when the noise had a standard deviation of 5. However, PNSR dropped quickly with higher noise, Overall, using LuminanceL1Loss demonstrates an improvement over a baseline of L1Loss. We used the reference implementation found at \href{https://github.com/majedelhelou/BUIFD}{https://github.com/majedelhelou/BUIFD}.

\section{Conclusion}

In this work, we have proposed LuminanceL1Loss, a novel loss function for image denoising and low-light enhancement that incorporates both color and luminance information. Our key finding is that adding an L1 penalty on the grayscale version of the predicted and target images helps capture perceptual brightness differences missed by common RGB losses like MSE.

We evaluated LuminanceL1Loss on three model architectures - Retinexformer, DnCNN, and BUIFD - across four datasets for image denoising and low-light enhancement. Our experiments demonstrate consistent improvements in PSNR and SSIM over baseline losses, with especially large gains on real low-light images where lighting realism is critical. Qualitative results also showcase more natural colors and lighting when using the proposed loss.

The results highlight the benefits of incorporating luminance signals directly into the optimization loss for image restoration tasks where lighting fidelity is important. By accounting for perceptual brightness, our LuminanceL1Loss allows models to better recreate the nuanced lighting patterns present in real-world images.

While our initial results are promising, there remain opportunities to build on this work. Future research directions include expanding the analysis to other tasks like dehazing, super-resolution and compression artifact removal where luminance may also be informative. Exploring different weighting schemes or learning an adaptive luminance weight may further improve performance. Overall, we believe the incorporation of explicit luminance-based losses could become an important technique in deep learning frameworks for low-level vision tasks. Models trained with losses aligned to human perceptual principles can continue to open new possibilities for image restoration.

\bibliography{sample}

\end{document}